\documentclass{bmvc2k}

\usepackage{algorithmic}
\usepackage{array}
\usepackage{textcomp}
\usepackage{stfloats}
\usepackage{url}
\usepackage{graphicx}
\usepackage{amsmath,amssymb,amsfonts}
\usepackage{multirow,multicol}
\usepackage{todonotes}
\usepackage{microtype}
\usepackage{overpic}
\usepackage{booktabs}
\usepackage{colortbl}
\usepackage{fontawesome5}
\usepackage{pgf-pie}
\usepackage{tikz}

\usepackage{capt-of}
\usepackage{lineno}

\usepackage{pgfplots}
\pgfplotsset{compat=1.18}
\usepackage{pgf-pie}
\definecolor{cMisc}{HTML}{B7B7B0}   

\definecolor{cEuler}{HTML}{D98B9F}  
\definecolor{cDino}{HTML}{8E8AD6}   
\definecolor{cRansac}{HTML}{6FC3B8} 
\definecolor{cDit}{HTML}{F0A860}    
\definecolor{cIcp}{HTML}{E8D26A}    
\definecolor{cDgedi}{HTML}{7CB66B}  
\definecolor{cUmap}{HTML}{E79CCB}   

\usepackage{geometry}
\title{Foundational feature fusion for conditional flow matching in 6D pose estimation}

\addauthor{Amir Hamza}{ahamza@fbk.eu}{1,2}
\addauthor{Davide Boscaini}{dboscaini@fbk.eu}{1}
\addauthor{Fabio Poiesi}{poiesi@fbk.eu}{1}

\addinstitution{
Fondazione Bruno Kessler\\
Trento, Italy
}
\addinstitution{
University of Trento\\
Trento, Italy
}

\runninghead{Hamza, Boscaini, Poiesi}{\acronym}

\begin{document}

\newcommand{\fabio}[1]{\todo[color=blue!20, inline, author=Fabio]{#1}}
\newcommand{\davide}[1]{\todo[color=yellow!20, inline, author=Davide]{#1}}
\newcommand{\amir}[1]{\todo[color=green!20, inline, author=Amir]{#1}}

\newcommand{\acronym}{FunFlow6D\xspace}

\def\eg{\emph{e.g}\bmvaOneDot}
\def\Eg{\emph{E.g}\bmvaOneDot}
\def\etal{\emph{et al}\bmvaOneDot}
\def\ie{\emph{i.e}\bmvaOneDot}

\newcommand{\warning}[1]{\textbf{\color{red!90}{#1}}}

\definecolor{myazure}{rgb}{0.8509,0.8980,0.9412}
\definecolor{mygreen}{RGB}{34,139,34}

\newcommand{\cmark}{\ding{51}}
\newcommand{\xmark}{\ding{55}}

\maketitle

\begin{abstract}
Conditional flow matching has enabled a step forward in object 6D pose estimation, achieving state-of-the-art performance by progressively denoising and registering object representations to observed scenes.
Existing methods require training task-specific encoders supervised on object-scene overlap and rely on trivial feature fusion strategies to resolve pose ambiguities.
We present \acronym, a novel flow matching-based formulation that leverages features from geometric and appearance foundation models for pose estimation, eliminating the need for task-specific encoder training.
We also introduce a cross attention-based fusion mechanism that dynamically combines geometric and appearance features to provide richer conditioning for the flow matching module.
Experiments on four datasets from the BOP benchmark show that \acronym outperforms the previous state of the art while reducing supervision requirements and memory overhead.
Extensive ablations validate the contribution of each proposed component.
Project website: \url{https://tev-fbk.github.io/FunFlow6D/}.
\end{abstract}

    
\section{Introduction}\label{sec:intro}

\begin{figure}[t]

    \vspace{1mm}
    \centering
    \begin{overpic}[trim=0 0 0 0, clip, width=1.0\linewidth]{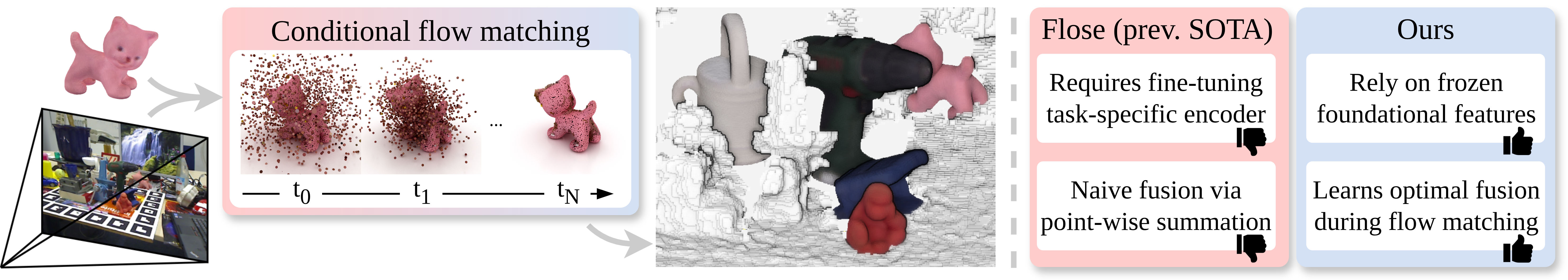}
    \end{overpic}
    \vspace{-7mm}

    \caption{
    \acronym estimates object 6D poses from input data (image + 3D model) via flow matching over multimodal foundational features (left).
    It advances over previous state-of-the-art on two fronts.
    First, it removes the need to train a task-specific encoder with object-scene overlap supervision (right, pink).
    Second, instead of relying on predefined fusion schemes such as point-wise summation, it learns how to optimally fuse multimodal features from frozen foundation models during flow matching (right, azure).
    As a result, \acronym achieves higher accuracy in a single training phase while requiring less supervision.
    }
    \label{fig:teaser}
\end{figure}

Manipulating and interacting with objects in the physical world requires accurate geometric localization.
Estimating an object's 6D pose (its position and orientation in the 3D environment) is key for spatial applications like augmented reality~\cite{augmentedreality}, robotic grasping~\cite{roboticgrasping}, and automated assembly~\cite{roboticassembly}.
These tasks demand accuracy under clutter~\cite{ycbv, icbin} and occlusions~\cite{lmo, nardon2025chip}.
In this work, we focus on \emph{model-based, instance-level 6D pose estimation} from RGBD data~\cite{ konig2020hybrid, lipson2022cir, liu2025gdrnpp}.
This setting assumes a prior 3D model of the object (CAD or reconstructed mesh) and explicit supervision over a closed set of predefined objects.
Traditional approaches~\cite{liu2025gdrnpp, wang2025hcceposebf, zebrapose, haugaard2022surfemb} cast this as a direct regression problem.
Recent work~\cite{hamza2026flose}, instead, formulates pose estimation as
iterative denoising via conditional flow matching (CFM)~\cite{sun2025rpf, Jin2025}, which needs fewer integration steps than diffusion~\cite{liu2023rectified}, trains more stably, and follows deterministic trajectories~\cite{lipman2023flow}.
The flow model learns a displacement field that transports Gaussian noise onto the object's 3D model, making CFM-based pose estimation inherently a registration problem. 
The flow routes each noisy scene point toward a target point on the CAD surface.
This routing is driven by per-point conditioning features, which enable point-to-point matching and must satisfy three properties:
(1) \emph{Geometric distinctiveness}: points on different surface neighborhoods must produce different features, otherwise the flow cannot route them apart.
(2) \emph{Local structure preservation}: nearby points must remain distinguishable even in a reduced-dimensional space, for tractability.
(3) \emph{Adaptive fusion}: conditioning must account for the fact that geometric and semantic cues are informative at different points.

State-of-the-art methods~\cite{hamza2026flose, sun2025rpf} lack these properties.
They condition the flow with an overlap-aware encoder~\cite{sun2025rpf, ptv3} trained to predict per-point visibility between the object's 3D model and the observed scene.
However, overlap may fail to encode local 3D structures.
We find that correspondences derived from overlap features yield only an average of 3.5 inliers on LM-O~\cite{lmo}, indicating that the vast majority of the geometric cues used to condition the flow matching stage are unreliable.
The overlap encoder also requires retraining for each new object, tying conditioning quality to per-instance supervision.
In contrast, handcrafted~\cite{fpfh, shot} and learned~\cite{fcgf, ppfnet, gedi, hamza2025dgedi} descriptors trained at scale are explicitly designed to be discriminative of local 3D structure and to generalize across shapes.
These descriptors~\cite{gedi, hamza2025dgedi} yield substantially higher inlier ratios than overlap-based features.
A 3D descriptor is therefore a more natural fit for the conditioning signal.
A separate limitation concerns feature dimensionality: both supervised~\cite{hamza2026flose} and unsupervised~\cite{caraffa2024freeze, caraffa2025accurate, ornek2024foundpose} methods reduce DINOv2~\cite{dinov2} features with PCA~\cite{pca} to match the dimensionality of the geometric branch. 
Being linear and unsupervised, PCA preserves the directions of highest global variance across the dataset, but this criterion is not necessarily aligned with what makes two nearby points distinguishable from one another.
As a result, directions that carry little variance overall but are locally discriminative, exactly the kind of fine-grained structure per-point conditioning depends on, can be discarded, creating an information bottleneck that limits how well nearby points can be told apart after reduction.
Finally, Flose~\cite{hamza2026flose} combines appearance and overlap features via point-level summation, which treats semantics and geometry as equally informative at every point. 
However, the informative modality varies across the object: textured regions rely on semantics, while geometrically distinct regions rely on shape.

We propose \acronym (\underline{f}o\underline{un}dational feature fusion for conditional \underline{flow} matching in \underline{6D} pose estimation), a CFM-based method that meets all three conditioning requirements. 
For geometric distinctiveness, we condition the flow on frozen dGeDi~\cite{hamza2025dgedi} features that are aware of local 3D structure and generalize well to unseen objects.
This raises the average number of inliers from 3.5 to 7.0 on LM-O (a $\sim$2$\times$ increase), without requiring any per-instance training.
For local structure preservation, we project frozen DINOv2 features with UMAP~\cite{Umap} instead of PCA.
UMAP is non-linear and explicitly preserves local neighborhoods, retaining task-relevant information that PCA may discard.
This raises the average number of inliers from 3.2 to 25.6 on LM-O (an $\sim$8$\times$ increase), at the same target dimensionality.
For adaptive fusion, we combine the two feature modalities with a learnable cross-attention gating module~\cite{gatedcrossattn}.
The module attends across modalities and learns per-point gates that determine how much each modality contributes.
This adapts the conditioning to the local structure of each object and enables richer cross-modal interaction than the point-wise summation used by Flose~\cite{hamza2026flose}.
We evaluate \acronym on four datasets from the BOP benchmark, covering diverse object categories, occlusion levels, textures, and clutter conditions. 
\acronym outperforms state-of-the-art methods while halving the number of trained parameters and removing the per-instance pre-training requirement.

In summary, our contributions are:
\begin{itemize}
    \itemsep 0em
    \item We condition flow matching for 6D pose estimation on robust correspondence signals extracted from a frozen, zero-shot geometric encoder, providing more discriminative conditioning features without requiring per-instance fine-tuning.
    \item We leverage UMAP to reduce the dimensionality of semantic features while preserving their local neighborhood structure, retaining task-relevant information that PCA's linear projection would discard.
    \item We introduce a gated-attention feature fusion module that adapts the flow matching conditioning to local object structure and promotes richer interaction between semantic and geometric cues.
\end{itemize}

%
%

\section{Related work}

\noindent \textbf{Instance-level 6D pose estimation methods} leverage explicit supervision on a closed set of known objects to estimate their 6D poses.
\emph{Model-based} approaches split into \emph{indirect} methods, which establish dense correspondences before solving for the pose, and \emph{direct} methods, which regress the transformation end-to-end.
\emph{Indirect methods} cast pose estimation as dense coordinate or feature regression~\cite{lmo, park2019pix2pose}, increasingly enriched by learned surface representations: SurfEmb~\cite{haugaard2022surfemb} learns continuous surface embeddings, ZebraPose~\cite{zebrapose} introduces a coarse-to-fine surface encoding, and HccePose(BF)~\cite{wang2025hcceposebf} extends this encoding to back-facing geometry for ultra-dense 2D--3D correspondences.
Complementary refinement~\cite{labbe2020cosypose, lipson2022cir, hu2022perspective} and hybrid~\cite{konig2020hybrid} pipelines improve robustness but remain bound by correspondence quality and degrade under occlusion and texture loss.
\emph{Direct methods} regress poses end-to-end on the $SE(3)$ manifold. 
GDR-Net~\cite{wang2021gdr} and GDRNPP~\cite{liu2025gdrnpp} couple geometric supervision with differentiable PnP, achieving state-of-the-art results on the BOP benchmark~\cite{hodan2024bop}. 
\emph{Generative formulations}, instead, model the full pose posterior, leveraging advances in denoising diffusion~\cite{Ho2020} and flow matching~\cite{lipman2023flow}.
They split into two families: those that learn a flow directly on the $SE(3)$ manifold, and those that transport points in $\mathbb{R}^3$ and recover the rigid transformation in closed form from the resulting correspondences.
SE(3)-PoseFlow~\cite{Jin2025} follows the former one, learning a flow on $SE(3)$ for uncertainty-aware manipulation.
Rectified Point Flow~\cite{sun2025rpf}, GARF~\cite{Li2025GARF}, and Register Any Point~\cite{Pan2025} belong to the latter, applying flow matching for registration and assembly in $\mathbb{R}^3$.
Closest to our work, Flose~\cite{hamza2026flose} casts instance-level 6D pose estimation as conditional flow matching that transports Gaussian noise onto the object's CAD surface, bypassing correspondence estimation and improving robustness to clutter and occlusion.
However, Flose conditions the flow on an overlap-aware encoder that requires retraining for every new object, reduces DINOv2 features with PCA, and fuses geometric and semantic cues by point-wise summation. 
In contrast, \acronym conditions flow matching on a frozen zero-shot 3D local descriptor to enable unseen-object generalization, introduces the use of UMAP~\cite{Umap} to preserve local neighborhood structure, and combines appearance and geometric cues through a learned cross-attention gating module.

\noindent \textbf{Per-point descriptors for 6D pose estimation.}
Correspondence-based and generative pose estimation methods recover the rigid transformation by matching or transporting the scene points to model points, respectively. 
Therefore, per-point descriptors must be discriminative over local surface neighborhoods and generalize to unseen objects without retraining. 
Foundation models trained at scale provide this signal, and divide into \emph{appearance} and \emph{geometric} encoders.
Appearance encoders are trained on large-scale RGB data with self-supervised objectives.
DINOv2~\cite{dinov2} outputs semantically-rich features, robust to illumination and viewpoint, and transferable to dense prediction without fine-tuning. 
It serves as a frozen backbone in both RGB~\cite{ornek2024foundpose} and RGBD~\cite{caraffa2024freeze, caraffa2025accurate} pose estimation pipelines.
Appearance features disambiguate textured regions but are uninformative about 3D structure: geometrically distinct points with similar texture share near-identical descriptors.
Geometric encoders are trained on 3D data to encode local surface structure. 
GeDi~\cite{gedi} learns generic and distinctive 3D descriptors with a PointNet++ backbone, achieving rotation invariance through local reference frames and generalizing across indoor, outdoor, and object-level scans. 
dGeDi~\cite{hamza2025dgedi} distills GeDi into a PTV3~\cite{ptv3} backbone that retains its discriminability by achieving higher inlier ratio than GeDi, while making inference faster.
The two modalities are complementary.
\acronym conditions flow matching jointly on DINOv2~\cite{dinov2} and dGeDi~\cite{hamza2025dgedi} features, and removes the training phase and overlap supervision required by the overlap-aware encoder.

\noindent \textbf{Feature fusion for 6D pose estimation} combines individual feature modalities into a unified representation.
Existing approaches can be divided into \emph{training-free} and \emph{training-based} methods.
\emph{Training-free fusion} combines the two modalities using closed-form operations.
FreeZe~\cite{caraffa2024freeze, caraffa2025accurate} concatenates semantic and geometric features along the channel dimension for training-free pose estimation.
Flose~\cite{hamza2026flose} uses point-wise summation to preserve the feature dimensionality required by the pre-trained backbone~\cite{sun2025rpf}.
Both fusion schemes are non-parameterized and treat the two modalities as equally informative at each point, with no mechanism to suppress unreliable cues or adapt the contribution of each modality independently.
\emph{Training-based fusion} relaxes these constraints by learning the optimal fusion strategy from data.
Standard cross-attention~\cite{attention} aggregates information across modalities but lacks an explicit per-point suppression mechanism.
Gated attention variants~\cite{gatedcrossattn} address this limitation by applying a head-specific sigmoid gate to the attention output, introducing input-dependent sparsity that selectively preserves or suppresses features on a per-point basis, with demonstrated gains in expressiveness and training stability.
\acronym fuses appearance- and geometry-aware features through a learned cross-attention gating module: cross-attention establishes per-point interactions between the two modalities, while a sigmoid gate determines how much each cue contributes at every point.
Unlike concatenation or summation, our fusion strategy adapts to local object structure; unlike plain cross-attention, it can suppress unreliable modalities on a per-point basis.

\section{Method}\label{sec:method}

\begin{figure}[t!]
    \centering
    \begin{overpic}[trim=0 0 6 0, clip, width=.99\linewidth]{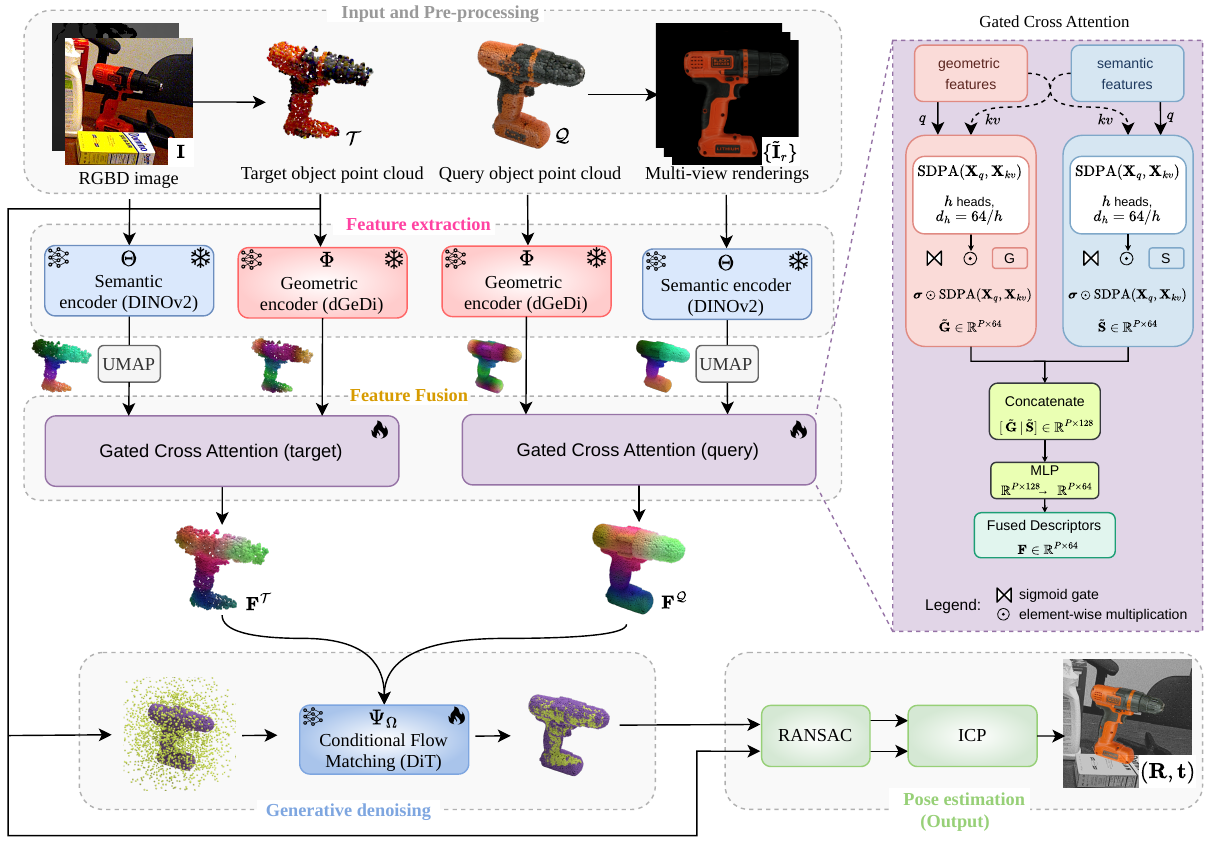}
    \end{overpic}

    \vspace{-2mm}
    \caption{
    Overview of \acronym.
    Given the canonical query point cloud $\mathcal{Q}$ and an RGBD image $\textbf{I}$ as input (top left), \acronym recovers the 6D object pose $(\textbf{R}, \textbf{t})$ (bottom right) through three stages: feature extraction, generative denoising, and pose estimation.
    Throughout, $\mathcal{Q}$ acts as a \emph{static anchor} and the flow operates only on the target $\mathcal{T}$, lifted from the masked depth of $\textbf{I}$.
    \textit{Feature extraction:} A frozen geometric encoder $\Phi$ (dGeDi) and a frozen semantic encoder $\Theta$ (DINOv2, reduced via UMAP) produce per-point descriptors that are fused by a bidirectional gated cross-attention module (right inset) into $\textbf{F}^\mathcal{Q}, \textbf{F}^\mathcal{T}$. 
    Colors encode feature similarity: corresponding regions across $\mathcal{Q}$ and $\mathcal{T}$ share similar colors.
    \textit{Generative denoising:} A conditional flow-matching network $\Psi_\Omega$ (DiT), conditioned on $\textbf{F}^\mathcal{Q}, \textbf{F}^\mathcal{T}$, learns a velocity field that canonicalizes $\mathcal{T}$ into $\hat{\mathcal{T}}$ while the anchor velocities are zeroed.
    \textit{Pose estimation:} The flow-induced correspondences between $\mathcal{T}$ and $\hat{\mathcal{T}}$ yield the camera-to-canonical transform via a RANSAC-based orthogonal Procrustes solver followed by ICP, and the pose $(\textbf{R}, \textbf{t})$.
    \faSnowflake{} denote frozen modules, while \faFire{} indicate trainable ones.
    }    
    \label{fig:diagram}
\end{figure}

\subsection{Problem formulation and Overview}

\noindent \textbf{Problem formulation.}
Let $\mathbf{I} \in \mathbb{R}^{H \times W \times 4}$ denote an RGBD observation of a given scene and $\mathcal{Q} \in \mathbb{R}^{N \times 3}$ the dense point cloud of a known \emph{query} object in its canonical reference frame.
Our primary objective is to recover the rigid transformation $\mathbf{T} \in \mathrm{SE}(3)$, parameterized by a rotation $\mathbf{R} \in \mathrm{SO}(3)$ and a translation $\mathbf{t} \in \mathbb{R}^3$, that maps $\mathcal{Q}$ from its canonical frame into the camera coordinate system.
To isolate the object of interest within the cluttered scene, we assume the availability of a binary segmentation mask $\mathbf{M} \in \{0, 1\}^{H \times W}$.
By applying $\mathbf{M}$ to the depth channel of $\mathbf{I}$ and back projecting the valid pixels into 3D space using the camera intrinsic matrix $\mathbf{K} \in \mathbb{R}^{3 \times 3}$, we obtain the \emph{target} point cloud $\mathcal{T} \in \mathbb{R}^{M \times 3}$, expressed in the camera reference frame.
Target point cloud is a partial, occluded, and noisy observation of the query object.
Formally, we seek $(\mathbf{R}, \mathbf{t})$ such that $\mathbf{R} \mathcal{Q} + \mathbf{t} \approx \mathcal{T}$.
To enable supervised alignment in canonical space, we further define $\mathcal{T}^r = \mathbf{R}^{\!\top}(\mathcal{T} - \mathbf{t}) \in \mathbb{R}^{M \times 3}$ as the target point cloud transported back into the canonical frame of $\mathcal{Q}$ using the ground-truth pose.

\noindent \textbf{Overview.}
We propose a three-stage pipeline driven by CFM shown in Fig.~\ref{fig:diagram}. 
Here $\mathcal{Q}$ acts as a \emph{static anchor} and is never transformed, the flow operates only on $\mathcal{T}$.
First, we process both $\mathcal{Q}$ and $\mathcal{T}$ to extract discriminative point-wise descriptors.
Appearance-aware semantic descriptors are obtained from frozen vision foundation model (VFM) and reduced via UMAP to preserve local neighborhood structures, while zero-shot geometric encoders provide complementary structural descriptors.
The two modalities are adaptively combined through a gated cross-attention network, providing fused representations $\mathbf{F}^\mathcal{Q}$ and $\mathbf{F}^\mathcal{T}$.
Second, conditioned on $(\mathbf{F}^\mathcal{Q}, \mathbf{F}^\mathcal{T})$, a flow-matching model deforms $\mathcal{T}$ into a canonicalized estimate $\hat{\mathcal{T}}$, predicting where each observed point lies in the canonical frame of $\mathcal{Q}$.
Third, the flow induces point-wise correspondence between $\mathcal{T}$ and $\hat{\mathcal{T}}$ from which we recover $\mathbf{T}' = (\mathbf{R}', \mathbf{t}') \in \mathrm{SE}(3)$ with $\mathbf{R}' \mathcal{T} + \mathbf{t}' \approx \hat{\mathcal{T}}$ via orthogonal Procrustes.
As $\mathbf{T}'$ is the camera-to-canonical mapping, the desired pose follows by inversion i.e. $\mathbf{T} = \mathbf{T}'^{-1}$.

\subsection{Feature extraction} 
We extract point-wise descriptors by jointly processing geometric features $\textbf{G}$ and semantic features $\textbf{S}$ for both the query $\mathcal{Q}$ and the target $\mathcal{T}$, before fusing them.

The geometric encoder $\Phi$ is a frozen, distilled multi-scale 3D local descriptor network~\cite{hamza2025dgedi} that, given a point cloud, produces a $D$-dimensional local geometric descriptor per point. 
It is distilled to encode information from two complementary spatial extents local neighborhoods covering $30\%$ and $40\%$ of the object's diameter in a single forward pass.
Applying $\Phi$ to $\mathcal{Q}$ and $\mathcal{T}$ yields multi-scale geometric features $\textbf{G}^\mathcal{Q} \in \mathbb{R}^{N \times D}$ and $\textbf{G}^\mathcal{T} \in \mathbb{R}^{M \times D}$.
Operating at two scales provides $\Phi$ with both fine-grained surface cues and broader contextual structure, yielding robustness to partiality and local noise in $\mathcal{T}$.

Geometric reasoning alone is insufficient to disambiguate objects exhibiting symmetries or sparse geometric detail. 
We therefore complement $\Phi$ with a semantic encoder $\Theta$ that lifts pixel-level features from a frozen VFM onto the 3D points of $\mathcal{Q}$ and $\mathcal{T}$. 
For the target, $\Theta$ is applied to a square crop of $\textbf{I}$ centered on the barycenter of $\mathcal{T}$, and the resulting pixel features are transferred to the points of $\mathcal{T}$ through the pixel-to-point correspondences induced by the depth channel, producing $\textbf{S}^\mathcal{T} \in \mathbb{R}^{M \times G}$, where $G$ denotes the VFM feature dimensionality.
For the query, since no RGB image is available at inference, we synthesize multi-view renderings $\{\widetilde{\textbf{I}}_r\}_{r=1}^{R}$ of its textured 3D model, apply $\Theta$ to each rendering, and back-project the resulting pixel features onto the points of $\mathcal{Q}$ using the renderer's camera intrinsics; whenever multiple pixels project onto the same 3D point, their features are aggregated by mean pooling across views. 
This rendering pass is performed once per object as an offline pre-processing step, so that at inference time the query features $\textbf{S}^\mathcal{Q} \in \mathbb{R}^{N \times G}$ are loaded from memory.

The dimensionality of $\textbf{S}^\mathcal{Q}$ and $\textbf{S}^\mathcal{T}$ is reduced from $G$ to $D$ via UMAP, which preserves the local manifold structure of the VFM embedding space more faithfully than linear projection based techniques.
Crucially, the UMAP embedding is fit only on $\textbf{S}^\mathcal{Q}$ and subsequently applied to project every instance of $\textbf{S}^\mathcal{T}$, ensuring that target and query semantic descriptors live on the same low-dimensional manifold.

\subsection{Feature fusion via gated cross-attention}
Rather than merging the two streams by point-wise addition, we let them interact through a bidirectional gated cross-attention block that adaptively weights, at every point, how much information each modality contributes.
The same module is applied independently to $\mathcal{Q}$ and $\mathcal{T}$; for simplicity we describe it for a generic object with $P$ points ($P{=}N$ for $\mathcal{Q}$, $P{=}M$ for $\mathcal{T}$). Let $\textbf{G}, \textbf{S} \in \mathbb{R}^{P \times D}$ denote its $L_2$-normalized geometric and semantic features.

The block contains two symmetric branches that exchange information across modalities. 
Each branch operates on an ordered pair of streams $(\textbf{X}_q, \textbf{X}_{kv})$. 
The first uses $(\textbf{X}_q, \textbf{X}_{kv}) = (\textbf{G}, \textbf{S})$, while the second swaps the roles to $(\textbf{S}, \textbf{G})$. 
Within each branch, queries are linearly projected from $\textbf{X}_q$, while keys and values are projected from $\textbf{X}_{kv}$, and multi-head scaled dot-product attention~\cite{attention} is computed over all $P$ points with $h$ heads of dimension $d_h = D/h$.
Following ~\cite{gatedcrossattn}, the SDPA output is modulated by a head-specific, element-wise sigmoid gate predicted from the query input:
\begin{equation}
\boldsymbol{\sigma} = \sigma(\textbf{X}_q \textbf{W}_g) \in \mathbb{R}^{P \times D}, \quad \mathrm{GAtt}(\textbf{X}_q, \textbf{X}_{kv}) = \boldsymbol{\sigma} \odot \mathrm{SDPA}(\textbf{X}_q, \textbf{X}_{kv}),
\end{equation}
where $\textbf{W}_g \in \mathbb{R}^{D \times D}$ holds the learnable gate parameters, $\sigma(\cdot)$ is the element-wise sigmoid, and $\odot$ the element wise product. 
The gated output is passed through a linear projection $\textbf{W}_O$, dropout, and a LayerNorm with a residual from $\textbf{X}_q$, yielding the refined streams $\tilde{\textbf{G}}, \tilde{\textbf{S}} \in \mathbb{R}^{P \times D}$ from the two branches respectively.
The biases of $\textbf{W}_g$ are zero-initialized so that $\boldsymbol{\sigma}$ starts at $0.5$ everywhere, letting both modalities contribute on equal footing at the beginning of training; the optimizer is then free to learn per-point, per-channel modality weights.

The two refined streams are concatenated along the channel axis and passed through a two-layer MLP with GELU activation that projects them back to $D$ dimensions, yielding the fused descriptor
\begin{equation}
\textbf{F} = \mathrm{norm}\!\big( \mathrm{LN}\big( \mathrm{MLP}([\,\tilde{\textbf{G}}\,|\,\tilde{\textbf{S}}\,]) \big) \big) \in \mathbb{R}^{P \times D},
\end{equation}
where $\mathrm{LN}(\cdot)$ denotes LayerNorm, $\mathrm{norm}(\cdot)$ denotes $L_2$ normalization along the channel axis, and $[\,\cdot\,|\,\cdot\,]$ denotes concatenation.
Applying the module to $\mathcal{Q}$ and $\mathcal{T}$ yields the per-point fused descriptors $\textbf{F}^{\mathcal{Q}} \in \mathbb{R}^{N \times D}$ and $\textbf{F}^{\mathcal{T}} \in \mathbb{R}^{M \times D}$, which condition the flow-matching model.

\subsection{Conditional flow matching}

Conditional flow matching learns a time dependent velocity field that transports samples from a simple noise distribution to a target data distribution along straight trajectories.
In our setup, we sample a noise state $\textbf{X}(0) \in \mathbb{R}^{2N \times 3}$, where each of the $2N$ points is drawn independently from a standard Gaussian $\mathcal{N}(0, I_3)$, and define the clean state as the concatenation of the anchor with the canonicalized target, $\textbf{X}(1) = [\,\mathcal{Q}\,;\,\mathcal{T}^r\,] \in \mathbb{R}^{2N \times 3}$, where $[\,\cdot\,;\,\cdot\,]$ denotes concatenation along the point axis.
The two states are linearly interpolated as
\begin{equation}
\textbf{X}(t) = (1-t)\,\textbf{X}(0) + t\,\textbf{X}(1), \qquad t \in [0,1],
\end{equation}
which yields a constant velocity $\textbf{X}(1) - \textbf{X}(0)$ along the trajectory.
A flow network $\Psi_\Omega$ is trained to regress this velocity conditioned on a signal $\textbf{C}$,
\begin{equation}
\Psi_\Omega\big(t, \textbf{X}(t)\,\big|\,\textbf{C}\big) \;\approx\; \textbf{X}(1) - \textbf{X}(0),
\end{equation}
by minimizing a mean square error function~\cite{sun2025rpf}, with $t$ sampled uniformly from $[0,1]$ and the pair $(\textbf{X}(0), \textbf{X}(1))$ drawn independently at each training step.

\vspace{1mm}
\noindent\textbf{Conditioning signal.}
The flow is conditioned on the fused appearance and geometry descriptors produced by the gated cross-attention module, augmented with a positional encoding,
\begin{equation}
\textbf{C} = [\,\textbf{F}\;|\;\textbf{P}\,] \in \mathbb{R}^{2N \times (D + d_P)},
\end{equation}
where $\textbf{F} = [\,\textbf{F}^\mathcal{Q}\,;\,\textbf{F}^\mathcal{T}\,]$ stacks the per-cloud fused descriptors along the point axis, $\textbf{P} \in \mathbb{R}^{2N \times d_P}$ is produced by a positional encoder $\Xi$, and $[\,\cdot\,|\,\cdot\,]$ denotes concatenation along the channel axis. 
For each point, $\Xi$ uses a $10$-dimensional input composed of its $3$D coordinates, its surface normal, the coordinates of its counterpart in the noise sample $\textbf{X}(0)$, and a binary flag distinguishing $\mathcal{Q}$ from $\mathcal{T}$. 
Sinusoidal embeddings~\cite{nerf} are applied to each geometric attribute independently, concatenated with learned point-wise features, and linearly projected to dimension $d_P$.

\vspace{1mm}
\noindent\textbf{Inference.}
At test time we recover the predicted position of points by integrating the learned velocity field forward through $K$ uniform Euler steps of size $\Delta t = 1/K$,
\begin{equation}
\hat{\textbf{X}}(t+\Delta t) = \hat{\textbf{X}}(t) + \Psi_\Omega\big(t, \hat{\textbf{X}}(t)\,\big|\,\textbf{C}\big)\,\Delta t,
\end{equation}
starting from a fresh Gaussian sample $\hat{\textbf{X}}(0)$, where each point is again drawn independently from $\mathcal{N}(0, I_3)$. Since $\mathcal{Q}$ acts as a static anchor, the velocity components corresponding to its points are zeroed out at every step, so that only the target portion of the state evolves. 
The target sub-block of $\hat{\textbf{X}}(1)$ is the canonicalized estimate $\hat{\mathcal{T}} \approx \mathcal{T}^r$.

\subsection{6D pose estimation}

The flow induces an explicit one-to-one correspondence between each point of the target $\mathcal{T}$, expressed in the camera frame, and its canonicalized counterpart in $\hat{\mathcal{T}}$. We exploit this correspondence to estimate the camera-to-canonical rigid transform $\mathbf{T}' = (\mathbf{R}', \mathbf{t}') \in \mathrm{SE}(3)$ such that $\mathbf{R}'\mathcal{T} + \mathbf{t}' \approx \hat{\mathcal{T}}$.
We obtain a coarse estimate $\mathbf{T}'_c$ by solving orthogonal Procrustes. 
At each iteration a small subset of correspondences is sampled, a candidate rigid transform is computed in closed form, and the hypothesis is scored by the number of inliers among the remaining pairs.
This rejects outliers induced by imperfect flow predictions, particularly in regions affected by heavy occlusion or symmetry ambiguity.
The coarse estimate is then refined through ICP aligning the transformed observation $\mathbf{R}'_c\mathcal{T} + \mathbf{t}'_c$ to the anchor $\mathcal{Q}$. 
The refinement proceeds in a coarse-to-fine manner over multiple decreasing distance thresholds. 
The final iteration produces the refined transform $\mathbf{T}'$, and the desired object pose follows by inversion $\mathbf{T} = \mathbf{T}'^{-1}$.

\section{Results}\label{sec:results}

\subsection{Experimental validation}

\noindent\textbf{Implementation details.}
We instantiate $\Phi$ with dGeDi~\cite{hamza2025dgedi} and $\Theta$ with DINOv2~\cite{dinov2}, and set the common descriptor dimensionality to $D = 64$, which matches the native output feature dimensionality of $\Phi$;
the DINOv2 features are reduced to the same dimensionality by UMAP.
The gated cross-attention block uses $h = 4$ heads, hence $d_h = 16$, with dropout $0.1$.
We fit UMAP with 30 neighbors, a minimum distance of $0.01$, and cosine similarity as metric.

\smallskip
\noindent \textbf{Data.}
We validate \acronym on four datasets from the BOP benchmark~\cite{lmo, tudl, ycbv, icbin}, spanning diverse object types (everyday vs. industrial), appearances (textured vs. textureless), and symmetries, as well as challenging environmental conditions such as heavy occlusion, dense clutter, and varying illumination.

\noindent \textbf{Metrics.}
To measure the distinctiveness of the features used to condition the flow matching process, we evaluate the \emph{average number of inliers} among object–scene correspondences.
We first establish correspondences between the object 3D model and the scene point cloud via mutual nearest-neighbor matching in the feature space.
We then transform the object 3D model using its ground-truth 6D pose to align it with the scene and compute the Euclidean distance between each matched point pair.
A correspondence is considered an inlier if this distance is below a threshold of 3\% of the object diameter.
To measure 6D pose accuracy, we use Average Recall (AR)~\cite{hodan2024bop}.
AR is the average of the recalls of three metrics: the Visible Surface Discrepancy (VSD), the Maximum Symmetry-Aware Surface Distance (MSSD), and the Maximum Symmetry-Aware Projection Distance (MSPD).
VSD measures pose error over the visible object surface, making it inherently invariant to symmetries.
MSSD captures the maximum 3D surface deviation under the set of valid symmetry transformations, reflecting suitability for grasping and manipulation.
MSPD measures the maximum 2D projection error, relevant for applications such as augmented reality.
For each of these three metrics, we compute recall over a range of error thresholds, obtaining $\text{AR}_{\text{VSD}}$, $\text{AR}_{\text{MSSD}}$, and $\text{AR}_{\text{MSPD}}$.
AR is then defined as $\text{AR} = (\text{AR}_{\text{VSD}} + \text{AR}_{\text{MSSD}} + \text{AR}_{\text{MSPD}})/3$.
To measure the method latency we compute the average inference time per image, measured in milliseconds (ms).
Latency is a practical constraint for the closed-loop downstream applications like robotic grasping and manipulation, which require pose updates at the sensor frame rate.


\subsection{Quantitative results}

\begin{figure}[t]
    \centering
    \begin{overpic}[trim=0 0 0 0, clip, width=0.98\columnwidth]{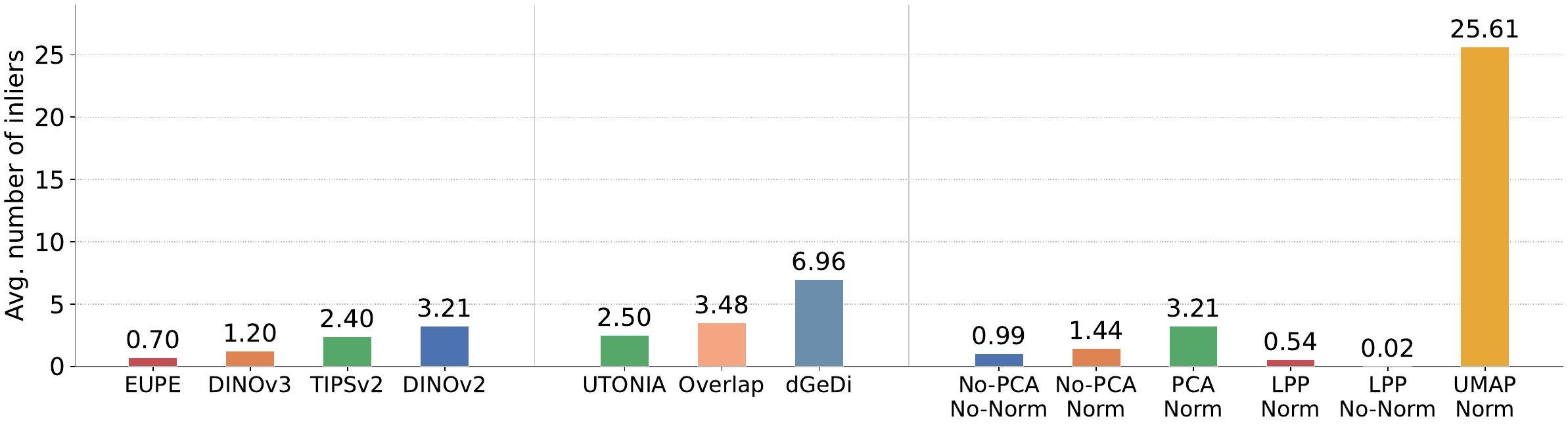}
    \end{overpic}

    \vspace{-3mm}
    \caption{
    Average number of inliers obtained with different appearance encoders (left), geometric encoders (center), and dimensionality reduction techniques (right).
    }
    \label{fig:inlier}
\end{figure}

\noindent\textbf{Feature discriminativity analysis.}
During flow matching, \acronym aligns points from a corrupted object representation with those of the observed scene.
Since this process is conditioned on point-level features, highly discriminative features are essential for establishing reliable object-scene correspondences.
To improve correspondence quality, Flose~\cite{hamza2026flose} fine-tunes the overlap-aware encoder introduced in RPF~\cite{sun2025rpf} on each BOP dataset~\cite{hodan2024bop}, a large-scale benchmark for 6D pose estimation. 
However, our experiments show that these overlap-aware features remain insufficiently discriminative despite the task-specific fine-tuning.
This motivated a thorough investigation of publicly available foundation models for geometric feature encoding.
Fig.~\ref{fig:inlier} summarizes the outcomes of this analysis in terms of average number of inliers on LM-O~\cite{lmo}.
It compares various geometric encoders (center), appearance encoders (left), and  dimensionality reduction techniques (right).
Among the geometric feature extractors, the overlap-aware encoder achieves an average number of inliers of 3.48, whereas dGeDi~\cite{hamza2025dgedi} reaches 6.96, yielding approximately a twofold improvement despite requiring no task-specific fine-tuning.
This is not observed for other foundation models, as the recent Utonia~\cite{utonia} encoder reaches only 2.5 inliers.

Among the appearance-aware feature extractors, DINOv2 outperforms both its follow up, DINOv3~\cite{dinov3}, and two recent approaches, EUPE~\cite{eupe} and TIPSv2~\cite{tipsv2}.
Specifically, DINOv2 has an average number of inliers of 3.21, surpassing TIPSv2 (2.40), DINOv3 (1.20), and EUPE (0.70). 
We attribute this gap to the nature of the supervision. DINOv2's self-distilled patch features encode dense, spatially-localized semantics that remain stable across viewpoint and illumination changes, whereas the more recent variants optimize for global or task-aligned objectives that smooth out the fine-grained local distinctive details our point-level conditioning depends on. 
Since correspondence routing operates per point rather than per image, an encoder that preserves intra-object feature variation is preferable to one that maximizes inter-image semantic consistency.
This motivates our choice of DINOv2 as an appearance encoder.

We discovered empirically that the specific technique used to reduce the dimensionality of the appearance encoder to match that of the geometric encoder plays an important role (Fig.~\ref{fig:inlier}, right).
Among the different dimensionality reduction techniques tested, UMAP yields by far the highest discriminativity, raising the average number of inliers to 25.61, an order of magnitude above the next best alternative. 
The linear baselines fall well short.
PCA reaches only 3.21, while applying no projection at all (No-PCA) gives 0.99 without normalization and 1.44 with it. 
Locality-preserving projection (LPP)~\cite{lpp}, despite being designed to retain neighborhood structure, collapses, dropping to 0.54 when normalized and to 0.02 without normalization.
Two observations follow.
First, the choice of projection matters more than the choice of whether to normalize.
The normalization toggle shifts the average number of inliers by a fraction of a point, whereas swapping PCA for UMAP improves it roughly eightfold.
Second, the failure of PCA and LPP confirms our hypothesis.
PCA preserves global variance but discards the local neighborhood structure that per-point conditioning relies on, and even an explicitly locality-aware linear method like LPP cannot recover the task-relevant manifold. 
UMAP's non-linear, neighborhood graph based embedding retains exactly these local relationships, keeping nearby points separable in the reduced space and thereby supplying the flow with a richer correspondence signal.
We therefore adopt DINOv2 features reduced via UMAP with normalization as our appearance conditioning.
This behavior is stable across runs: repeating the UMAP fit with five random
seeds yields $25.4 \pm 0.5$ inliers on LM-O, against the $25.6$ obtained with our fixed seed.

\begin{table}[t!]
\centering
\caption{
6D pose accuracy on the BOP benchmark, evaluated in terms of AR.
}
\label{tab:quant}

\smallskip
\resizebox{\linewidth}{!}{%
\begin{tabular}{rlrrrrrrrr}
    \toprule
    & Method & Models & Params & LM-O & TUD-L & IC-BIN & YCB-V & Avg \\
    \toprule

    \color{gray} \footnotesize 1 & HccePose(BF)~\cite{wang2025hcceposebf} & 34 & 41.5M & 80.5 & 94.4 & 72.4 & 91.1 & 84.6 \\

    \color{gray} \footnotesize 2 & GDRNPP (BOP23)~\cite{liu2025gdrnpp} & 4 & 106.8M & 79.4 & 96.4 & 73.7 & 92.8 & 85.6 \\

    \color{gray} \footnotesize 3 & Flose~\cite{hamza2026flose} & 4 & 89.2M & 86.1 & 98.8 & 74.8 & 92.9 & 88.2 \\

    \rowcolor{myazure}
    \color{gray} \footnotesize 4 & \acronym (ours) & 4 & 43.1M & \textbf{87.1} & \textbf{98.9} & \textbf{75.3} & \textbf{93.6} & \textbf{88.7} \\

    \rowcolor{myazure}
    \color{gray} \footnotesize 5 & Improvement wrt row 3 & - & - & \textcolor{mygreen}{+1.0} & \textcolor{mygreen}{+0.1} & \textcolor{mygreen}{+0.5} & \textcolor{mygreen}{+0.7} & \textcolor{mygreen}{+0.5} \\

    \bottomrule
\end{tabular}
}
\end{table}

\smallskip
\noindent\textbf{6D pose estimation accuracy.}
Tab.~\ref{tab:quant} reports a quantitative evaluation of 6D pose estimation accuracy in terms of AR.
To keep the table compact, we compare \acronym against the strongest indirect method (row~1, HccePose(BF)~\cite{wang2025hcceposebf}), the strongest direct method (row~2, GDRNPP~\cite{liu2025gdrnpp}), and the strongest flow matching-based approach (row~3, Flose~\cite{hamza2026flose}).
The ``Models'' column shows the number of neural networks trained by each method.
HccePose(BF) trains one network for each object (34 models in total),
whereas all other methods train a single network per dataset, requiring about 9$\times$ fewer models.
The ``Params'' column shows the number of trainable parameters per model.
\acronym is the most parameter-efficient approach, requiring approximately 9$\times$ fewer parameters than HccePose(BF) and less than half the parameters of Flose.
The remaining columns report AR on LM-O, TUD-L, IC-BIN, YCB-V, and their average.
\acronym consistently outperforms all competing methods on all four datasets. Row~5 shows the absolute improvement over Flose, which is both the closest approach to ours and the best performing competitor overall.
Despite requiring substantially fewer trainable parameters and significantly weaker supervision, \acronym achieves consistent gains across all datasets.
Interestingly, the largest improvement is observed on LM-O, a dataset characterized by severe occlusions, suggesting that \acronym is particularly effective in scenarios where robust geometric reasoning is required.
Notably, these gains are achieved while conditioning on frozen encoders and without any per-instance training or finetuning, unlike all competing approaches.

\begin{figure}[t!]
    \centering
    \footnotesize
    \begin{minipage}{0.48\columnwidth}
        \centering
        \begin{minipage}{0.04\columnwidth}
        \end{minipage}%
        \begin{minipage}{0.32\columnwidth}
        \centering Input
        \end{minipage}%
        \begin{minipage}{0.32\columnwidth}
        \centering Flose~\cite{hamza2026flose}
        \end{minipage}%
        \begin{minipage}{0.32\columnwidth}
        \centering \acronym
        
        \end{minipage}

        \begin{minipage}{0.04\columnwidth}
        \rotatebox{90}{\footnotesize Duck}
        \end{minipage}%
        \begin{minipage}{0.96\columnwidth}
        \centering
        \includegraphics[width=\columnwidth]{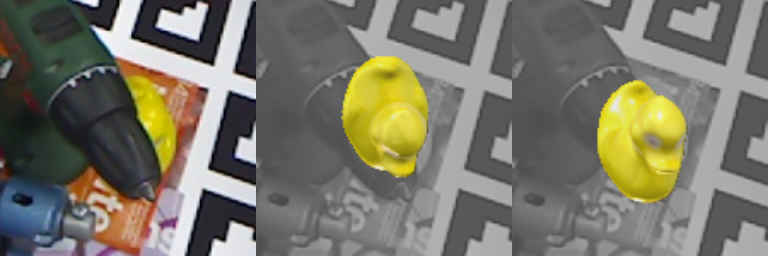}
        \end{minipage}

        \begin{minipage}{0.04\columnwidth}
        \rotatebox{90}{\footnotesize Ape}
        \end{minipage}%
        \begin{minipage}{0.96\columnwidth}
        \centering
        \includegraphics[width=\columnwidth]{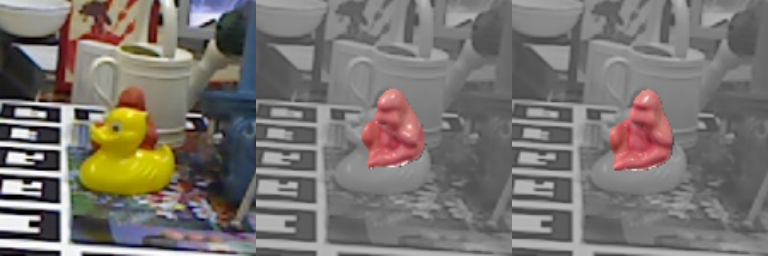}
        \end{minipage}

        \begin{minipage}{0.04\columnwidth}
        \rotatebox{90}{\footnotesize Glue bottle}
        \end{minipage}%
        \begin{minipage}{0.96\columnwidth}
        \centering
        \includegraphics[width=\columnwidth]{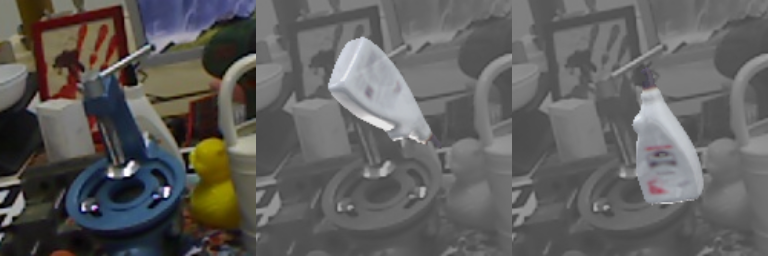}
        \end{minipage}

        \begin{minipage}{0.04\columnwidth}
        \rotatebox{90}{\footnotesize Cat}
        \end{minipage}%
        \begin{minipage}{0.96\columnwidth}
        \centering
        \includegraphics[width=\columnwidth]{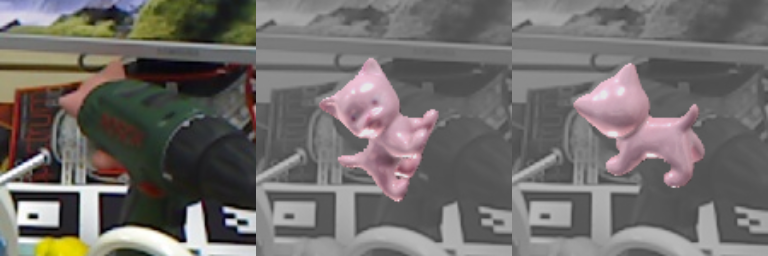}
        \end{minipage}

        \begin{minipage}{0.04\columnwidth}
        \rotatebox{90}{\footnotesize Egg box}
        \end{minipage}%
        \begin{minipage}{0.96\columnwidth}
        \centering
        \includegraphics[width=\columnwidth]{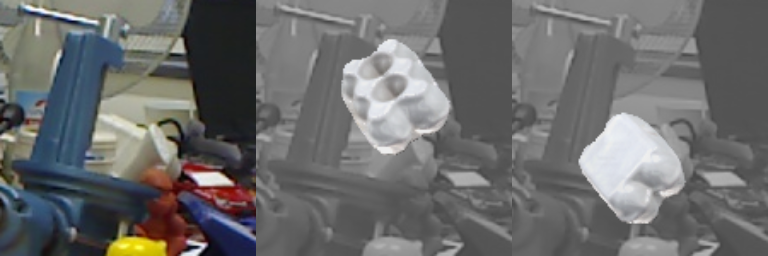}
        \end{minipage}

        \begin{minipage}{0.04\columnwidth}
        \rotatebox{90}{\footnotesize Hole punch}
        \end{minipage}%
        \begin{minipage}{0.96\columnwidth}
        \centering
        \includegraphics[width=\columnwidth]{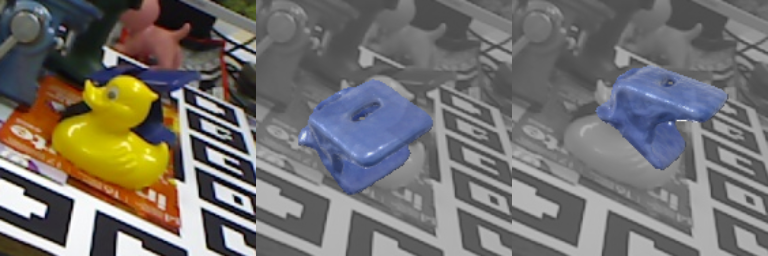}
        \end{minipage}

        \begin{minipage}{0.04\columnwidth}
        \rotatebox{90}{\footnotesize Ape}
        \end{minipage}%
        \begin{minipage}{0.96\columnwidth}
        \centering
        \includegraphics[width=\columnwidth]{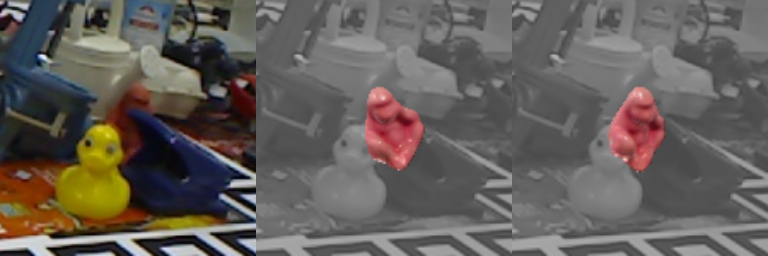}
        \end{minipage}

        \vspace{0.5mm}
        \centerline{\footnotesize (a) LM-O}
    \end{minipage}%
    \hfill
    \begin{minipage}{0.48\columnwidth}
        \centering
        \begin{minipage}{0.04\columnwidth}
        \end{minipage}%
        \begin{minipage}{0.32\columnwidth}
        \centering Input
        \end{minipage}%
        \begin{minipage}{0.32\columnwidth}
        \centering Flose~\cite{hamza2026flose}
        \end{minipage}%
        \begin{minipage}{0.32\columnwidth}
        \centering \acronym
        \end{minipage}

        \begin{minipage}{0.04\columnwidth}
        \rotatebox{90}{\footnotesize Tomato soup can}
        \end{minipage}%
        \begin{minipage}{0.96\columnwidth}
        \centering
        \includegraphics[width=\columnwidth]{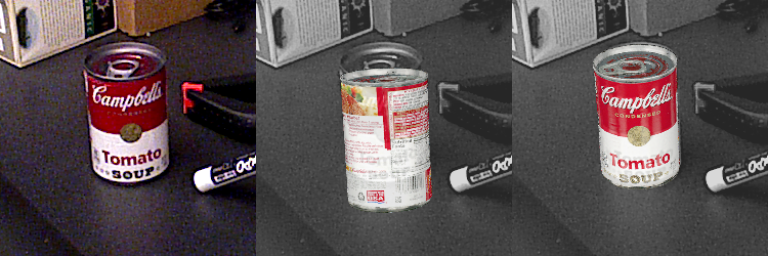}
        \end{minipage}

        \begin{minipage}{0.04\columnwidth}
        \rotatebox{90}{\footnotesize Tuna fish can}
        \end{minipage}%
        \begin{minipage}{0.96\columnwidth}
        \centering
        \includegraphics[width=\columnwidth]{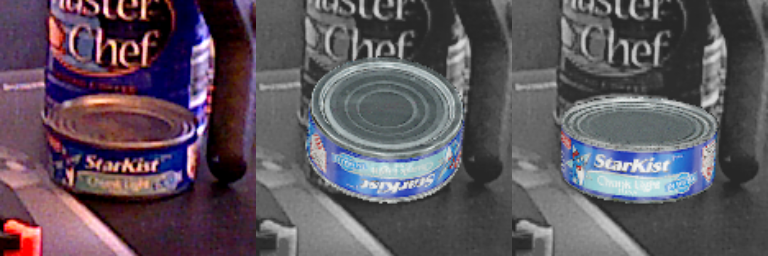}
        \end{minipage}

        \begin{minipage}{0.04\columnwidth}
        \rotatebox{90}{\footnotesize Pudding box}
        \end{minipage}%
        \begin{minipage}{0.96\columnwidth}
        \centering
        \includegraphics[width=\columnwidth]{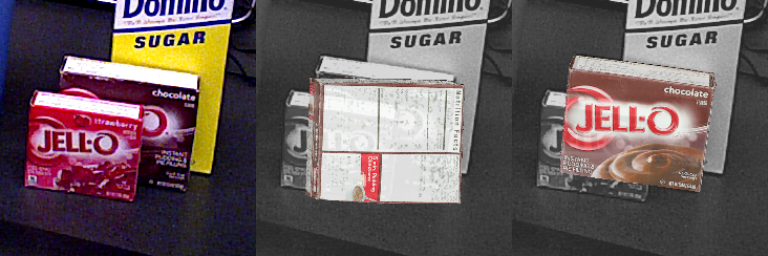}
        \end{minipage}

        \begin{minipage}{0.04\columnwidth}
        \rotatebox{90}{\footnotesize Clamp}
        \end{minipage}%
        \begin{minipage}{0.96\columnwidth}
        \centering
        \includegraphics[width=\columnwidth]{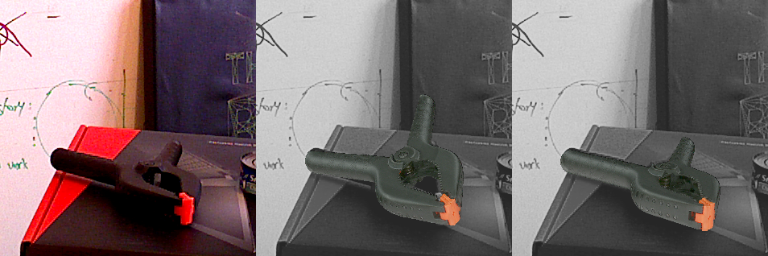}
        \end{minipage}

        \begin{minipage}{0.04\columnwidth}
        \rotatebox{90}{\footnotesize Master Chef can}
        \end{minipage}%
        \begin{minipage}{0.96\columnwidth}
        \centering
        \includegraphics[width=\columnwidth]{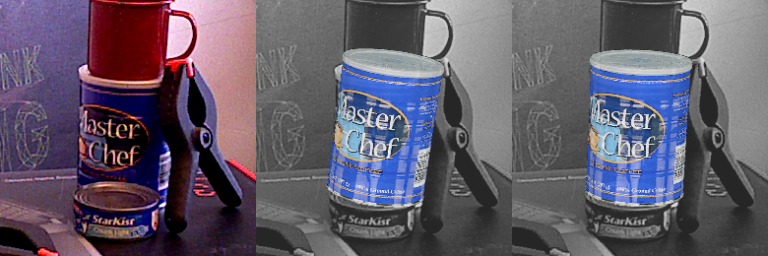}
        \end{minipage}

        \begin{minipage}{0.04\columnwidth}
        \rotatebox{90}{\footnotesize Scissor}
        \end{minipage}%
        \begin{minipage}{0.96\columnwidth}
        \centering
        \includegraphics[width=\columnwidth]{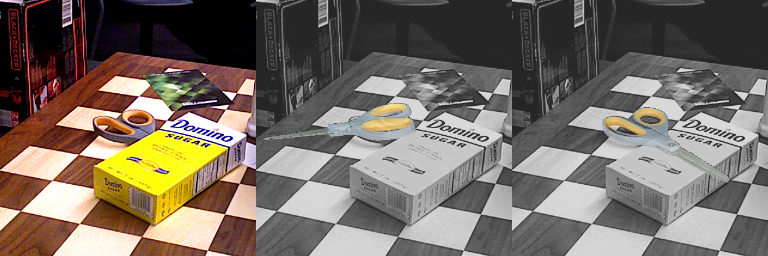}
        \end{minipage}

        \begin{minipage}{0.04\columnwidth}
        \rotatebox{90}{\footnotesize Block}
        \end{minipage}%
        \begin{minipage}{0.96\columnwidth}
        \centering
        \includegraphics[width=\columnwidth]{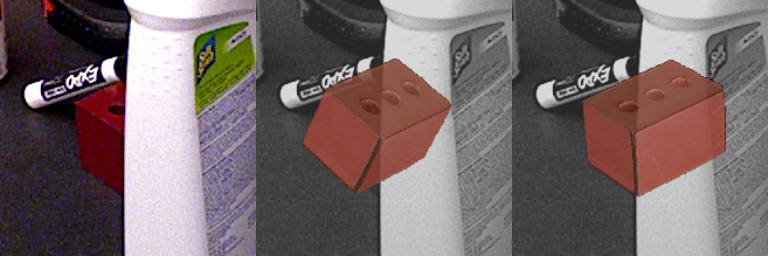}
        \end{minipage}

        \vspace{0.5mm}
        \centerline{\footnotesize (b) YCB-V}
    \end{minipage}
    \vspace{2mm}
    \caption{
    \acronym (right) vs.~Flose~\cite{hamza2026flose} (center) on LM-O (a) and YCB-V (b). \acronym predicts more accurate poses under severe occlusions and symmetry ambiguities.
    }
    \label{fig:qual}
\end{figure}

\subsection{Qualitative results}
Fig.~\ref{fig:qual} shows qualitative comparisons on LM-O and YCB-V between \acronym and Flose~\cite{hamza2026flose}.
Each image shows a rendering of the object's 3D model in the predicted pose, overlaid on a grayscale version of input image.
Prediction accuracy can be assessed by the alignment between the colored rendering and the visible portion of the object in the grayscale background.
Rows 1--7 (LM-O) show objects under severe occlusion and clutter: a duck occluded by a drill, an ape severely occluded by a duck, a glue bottle partially hidden by a mechanical part and a cat highly occluded by the drill, a white egg box heavily occluded by multiple objects in the front, a smooth textured but geometrically distinct hole punch occluded by the duck, another instance of less occluded ape.
In these cases, the overlap encoder in Flose provides unreliable point-to-point cues, whereas the discriminative geometric descriptor in \acronym lets the flow route scene points to the model surface more accurately.
Rows 1--7 (YCB-V) show objects where appearance is the dominant cue: a tomato soup can, a tuna fish can, a richly textured pudding box, a geometrically distinct clamp, a symmetric master chef can, a pair of scissors occluded by a sugar box, and a block heavily occluded by a white bottle.
Flose misaligns the tomato soup can near its top rim and flips both the tuna fish can and the pudding box, errors caused by relying on near-symmetric geometry where the only discriminative cues are textural.
On the master chef can, Flose misaligns the top of the can, again driven by its rotational near-symmetry.
For the scissors, only the knuckle ends are clearly visible, and while both methods align that region, Flose places the pointed blades on the wrong side, producing a flipped pose; \acronym recovers the correct orientation.
On the block, Flose latches onto a single edge and aligns to it, leaving the rest of the object offset.
In \acronym, the brand text and logos supply distinctive semantic features, and the learned gating module up-weights the DINOv2 descriptors over geometry, resolving the symmetry and alignment ambiguities that the geometric branch alone cannot.


\subsection{Inference time breakdown and ablation study}

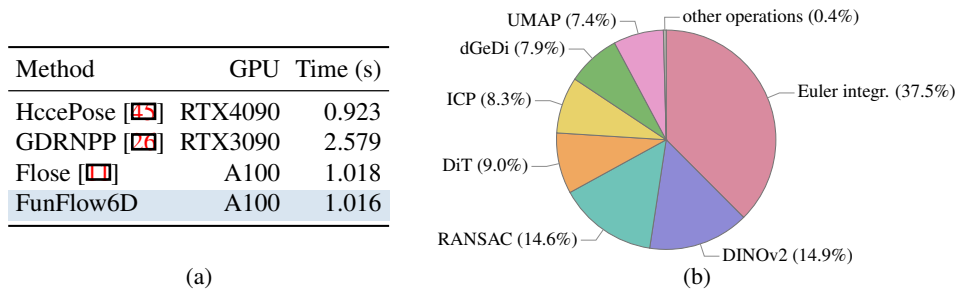
\begin{figure}[t]
\centering

%
%
\begin{minipage}[b]{0.39\textwidth}
  \centering
  \tabcolsep 3pt
  \resizebox{\linewidth}{!}{%
  \begin{tabular}{lrr}
    \toprule
    Method   & GPU     & Time (s) \\
    \midrule
    HccePose~\cite{wang2025hcceposebf} & RTX4090 & 0.923 \\
    GDRNPP~\cite{liu2025gdrnpp}         & RTX3090 & 2.579 \\
    Flose~\cite{hamza2026flose}        & A100    & 1.018 \\
    \rowcolor{myazure}
    \acronym                           & A100    & 1.016 \\
    \bottomrule
  \end{tabular}}
  \par\vspace{5mm}    
  {\small (a)}
\end{minipage}
\hfill
%
%
\begin{minipage}[b]{0.59\textwidth}
  \centering
  \begin{tikzpicture}[font=\scriptsize]
    \def\Rad{1.45}
    \def\Lab{1.62}
    \xdef\startang{90}
    
    \foreach \pct/\col/\dy/\side/\lr/\lbl in {%
        37.52/cEuler/0/0/1.62/Euler integr. (37.5\%),%
        14.89/cDino/0/0/1.62/DINOv2 (14.9\%),%
        14.61/cRansac/0/0/1.62/RANSAC (14.6\%),%
        8.98/cDit/0/0/1.62/DiT (9.0\%),%
        8.33/cIcp/0/0/1.62/ICP (8.3\%),%
        7.87/cDgedi/0/0/1.62/dGeDi (7.9\%),%
        7.39/cUmap/0/0/1.62/UMAP (7.4\%),%
        0.42/cMisc/0/1/2.05/other operations (0.4\%)}
    {%
      \pgfmathsetmacro{\sweep}{\pct*3.6}%
      \pgfmathsetmacro{\stopang}{\startang-\sweep}%
      \pgfmathsetmacro{\midang}{\startang-\sweep/2}%
      \fill[fill=\col,draw=black!55,line width=0.2pt]
        (0,0) -- (\startang:\Rad)
        arc[start angle=\startang,end angle=\stopang,radius=\Rad] -- cycle;
        \pgfmathparse{(\side==1 || (\side==0 && cos(\midang)>0)) ? "west" : "east"}%
      \edef\anch{\pgfmathresult}%
      \pgfmathsetmacro{\dx}{(\side==1 || (\side==0 && cos(\midang)>0)) ? 0.20 : -0.20}%
      \draw[black!45,line width=0.2pt]
        (\midang:\Rad) -- (\midang:\Lab) -- ++(\dx,0);
      \node[anchor=\anch,inner sep=1pt,yshift=\dy mm]
        at ([xshift=\dx cm]\midang:\Lab) {\lbl};
      \xdef\startang{\stopang}%
    }%
  \end{tikzpicture}
  \par\vspace{-2mm}
  {\small (b)}
\end{minipage}

\vspace{-3mm}
\caption{
Inference time analysis on LM-O.
(a) Average per-image runtime against competing methods. Object instance segmentation is excluded for all methods.
(b) Per-image breakdown of \acronym. The Euler integration (50 steps) dominates the pipeline.
}
\label{fig:timing}
\end{figure}

\noindent \textbf{Inference time breakdown.}
We provide here the per-image inference time breakdown for LM-O (NVIDIA A100 SXM4 64GB, 32-core Intel Xeon CPU), which contains 8 objects across 200 images and on average features all of them per image.
Timings are measured per image.
For segmentation, we use precomputed ZebraPose~\cite{zebrapose} masks (NVIDIA RTX 2080Ti, Intel Xeon E-2146G CPU @ 3.50GHz) available on the BOP leaderboard, taking 0.08\,s per image. 
For the rest of the pipeline, i.e. the geometric encoder, semantic encoder, and flow model are batched according to the number of segmented instances in the image, so that all instances of an image are processed in a single forward pass.
For the dimensionality reduction we adopt the GPU implementation of UMAP from TorchDR~\cite{Van_Assel_TorchDR}.
Fig.~\ref{fig:timing}(a) compares \acronym average per-image runtime against the competing methods, where HccePose and GDRNPP's runtimes are taken from the public BOP leaderboard~\cite{bop_leaderboard}.
\acronym runtime is comparable with that of Flose, evaluated on the same GPU type as ours, slightly slower than HccePose, and faster than GDRNPP.
The detailed cost per image for \acronym is as follows: DINOv2 forward pass takes 151.29 ms (14.89\%), UMAP projection takes 75.10 ms (7.39\%), the dGeDi geometric encoding takes 79.90 ms (7.87\%), the DiT (flow model) forward pass takes 91.21 ms (8.98\%), Euler integration at 50 steps takes 381.11 ms (37.52\%), RANSAC registration (1k iters, Open3D) takes 148.45 ms (14.61\%), ICP refinement (3k iters, Open3D) takes 84.59 ms (8.33\%), and other operations (lifting point clouds, downsampling, and outlier removal) take 4.25 ms (0.42\%), resulting in 1015.90 ms per image.
The Euler integration accounts for the largest share of the total time (37.5\%), which is inherent to iterative generative approaches. 
It is followed by the DINOv2 forward pass (14.9\%) and RANSAC registration (14.6\%). 
The full breakdown is summarized in Fig.~\ref{fig:timing}(b).

\begin{table}[t!]
\centering
\caption{Ablation study on LM-O~\cite{lmo}.
Key: \faFire{} = trained, \faSnowflake{} = frozen, - = not used.
The default configuration of \acronym is \colorbox{myazure}{highlighted}.
}
\label{tab:ablation}

\smallskip
\resizebox{\linewidth}{!}{%
\begin{tabular}{rllllllr}
    \toprule
    & Geom. Enc. & App. Enc. & Dim. Red. & Feat. fusion & Pose estim. & Pose refin. & AR \\
    \toprule
    \color{gray} \footnotesize 1 & \faFire{} overlap-aware & - & - & - & SVD & ICP & 83.5 \\
    \color{gray} \footnotesize 2 & \faFire{} overlap-aware & \faSnowflake{} DINOv2 & PCA & \faSnowflake{} Point-level sum & RANSAC & ICP & 86.1 \\
    \color{gray} \footnotesize 3 & \faSnowflake{} dGeDi & \faSnowflake{} DINOv2 & PCA & \faSnowflake{} Point-level sum & RANSAC & ICP & 81.8 \\
    \color{gray} \footnotesize 4 & \faSnowflake{} dGeDi & \faSnowflake{} DINOv2 & UMAP & \faSnowflake{} Point-level sum & RANSAC & ICP & 83.6 \\
    \color{gray} \footnotesize 5 & \faSnowflake{} dGeDi & \faSnowflake{} DINOv2 & UMAP & \faSnowflake{} No attention & RANSAC & ICP & 85.7 \\
    \color{gray} \footnotesize 6 & \faSnowflake{} dGeDi & \faSnowflake{} DINOv2 & UMAP & \faFire{} Cross attention & RANSAC & ICP & 86.3 \\
    \color{gray} \footnotesize 7 & \faSnowflake{} dGeDi & \faSnowflake{} DINOv2 & UMAP & \faFire{} Gated attention & RANSAC & - & 85.8 \\
    \rowcolor{myazure} \color{gray} \footnotesize 8 & \faSnowflake{} dGeDi & \faSnowflake{} DINOv2 & UMAP & \faFire{} Gated attention & RANSAC & ICP & \textbf{87.1} \\
    \bottomrule
\end{tabular}
}

\end{table}
\smallskip
\noindent\textbf{Ablation study.}
Tab.~\ref{tab:ablation} dissects each component of \acronym on LM-O~\cite{lmo}, isolating the contribution of geometric encoder, dimensionality reduction of semantic features, fusion strategy, and pose refinement stage. 
We start from an overlap-aware baseline and progressively replace each module with our proposed design.  \noindent\textit{Geometric conditioning.}
Rows 1--2 establish the baseline, which conditions the flow on a pre-trained overlap-aware encoder~\cite{sun2025rpf}. 
Adding frozen DINOv2~\cite{dinov2} features reduced with PCA and fused by point-level summation raises AR from 83.5 to 86.1, confirming that semantic cues complement geometry. 
Replacing the overlap-aware encoder with \emph{frozen} zero-shot dGeDi descriptor~\cite{hamza2025dgedi} under the same PCA and point-level sum scheme yields 81.8 AR.
This configuration remains competitive while using no instance specific supervision for feature extraction. \noindent\textit{Semantic dimensionality reduction.}
Rows 3--4 isolate the effect of the reduction technique while keeping every other component fixed. 
Substituting PCA with UMAP~\cite{Umap} improves AR from 81.8 to 83.6 (+1.8). 
UMAP is non-linear and explicitly preserves local neighborhoods, it retains task-relevant directions that PCA's linear, variance-driven projection drops below threshold. 
This is consistent with the correspondence-level evidence reported in our analysis, where DINOv2-UMAP raises the average number of inliers from 3 to 25 at the same target dimensionality.
\noindent\textit{Fusion strategy.} Rows 4--6 replace the static point-level summation with learned fusion.
Our module consists of gated cross-attention followed by concatenation and MLP adaptation, and we isolate each of these in turn.
Removing the attention entirely and retaining only the concatenation and MLP adaptation (row 5) already raises AR from 83.6 to 85.7, showing that a parameterized combination outperforms an equally weighted sum.
Restoring the cross-modal interaction with standard, ungated cross-attention (row 6) adds a further $+0.6$, as every point can now attend to the complementary modality rather than being reweighted channel-wise.
Adding the sigmoid gate (row 8) gives the best result, since each point can now suppress the unreliable modality instead of only mixing the two.
This progression quantifies the limitation of treating semantics and geometry as equally informative. 
Textured regions benefit from appearance while geometrically salient regions rely on shape, and the per-point gates learn to route each cue accordingly.
Even without any pose refinement, gated attention (row 7) reaches 85.8 AR, exceeding the summation variant that uses ICP refinement (row 4). \noindent\textit{Pose refinement.}
Rows 7--8 sweep the refinement stage on top of the full conditioning pipeline.
The coarse RANSAC estimate alone yields 85.8 AR (row 7). 
ICP recovers an additional +1.3 AR (row 8) by correcting residual alignment errors.
The proposed pipeline (row 8) attains 87.1 AR, a +3.6 improvement over the overlap-aware baseline (row 1), while eliminating the pre-training requirement. 
The three proposed components are complementary. 
Each contributes a measurable gain and their combination accounts for the full margin over the baseline.



\section{Conclusion}\label{sec:conclusion}

We presented \acronym, the first conditional flow matching method that conditions the flow matching process via fusion of appearance and geometric foundational features to
estimate the 6D pose of known objects from RGBD observations of a scene.
Compared to previous state-of-the-art methods, \acronym conditions the generative denoising process on multimodal foundational features, employing a gated attention-based fusion strategy to merge them.
Experiments demonstrate that \acronym outperforms state-of-the-art competitors on four datasets from the BOP benchmark, achieving superior robustness to symmetries and occlusions while requiring a simplified training recipe and weaker supervision signals.

\smallskip
\noindent\textbf{Limitations and future work.}
(1) Our pipeline relies on an external module to estimate 6D poses via RANSAC, as the output of the flow matching process is not guaranteed to be a valid rigid transformation. Enforcing rigidity constraints during denoising would make the approach fully end-to-end and more efficient.
(2) During inference, \acronym requires multiple denoising steps, which increases the latency of the method. In the future, we will explore single-step flow matching alternatives to improve the latency.
(3) \acronym relies on segmentation masks, this makes performance contingent on segmentation quality and introduces an additional pipeline stage that is not jointly optimized.
A future direction can be removing the localization stage and predicting the flow directly at scene level point clouds.


\bibliography{main}

\end{document}